\documentclass[11pt,a4paper]{article}

\usepackage[utf8]{inputenc}
\usepackage[T1]{fontenc}
\usepackage{lmodern}
\usepackage[margin=25mm]{geometry}
\usepackage{amsmath,amssymb,amsthm}
\usepackage{graphicx}
\usepackage{svg}            
\usepackage{booktabs}
\usepackage{xcolor}
\usepackage[hidelinks]{hyperref}
\usepackage{microtype}
\usepackage{authblk}
\usepackage{caption}
\usepackage{titlesec}
\usepackage{enumitem}
\usepackage{lineno}

\titleformat{\section}{\normalfont\large\bfseries}{\thesection}{0.6em}{}
\titleformat{\subsection}{\normalfont\normalsize\bfseries}{\thesubsection}{0.5em}{}

\newcommand{\imgsim}{\textsc{img2sim}}
\newcommand{\vv}{V\&V}

\title{\textbf{Instrumented data for causal\\
scientific machine learning}\\[0.4em]
}

\author[1]{Daniel N.~Wilke\thanks{Corresponding author: \texttt{daniel.wilke@wits.ac.za}}}
\affil[1]{School of Mechanical, Industrial and Aeronautical Engineering, University of the Witwatersrand, Johannesburg, South Africa}

\date{}

\begin{document}
\maketitle

\begin{abstract}
\noindent
Scientific machine learning is limited less by model size than by the
data it is trained on. Observational data records what happened but
not why; template synthetic data has a known generating process but
only for the simulator's template, not the case a user faces. We
argue a third option is now operationally feasible: \emph{instrumented
data}, in which every datum carries the mechanistic model that
produced it, an explicit uncertainty over that model, and an
executable family of counterfactuals. Verification-and-validation
(V\&V) instrumented image-to-simulation pipelines are one realisation:
a sensor observation becomes a fully specified, solver-backed
simulation with explicit, editable parameters and a propagated
aleatoric/epistemic uncertainty. The substrate is case-specific,
mechanistically supervised, and supports causal interventions through
Pearl's $\mathrm{do}$-operator. Near-term consequences for validation,
auditing, and surrogate training span computational biology, climate,
materials, fluid mechanics, and medical imaging; a longer-term,
falsifiable implication concerns foundation models for scientific
reasoning.
\end{abstract}

\section{A data problem cutting across computational science}
\label{sec:intro}

Most machine learning today is trained on one of two kinds of data.
\emph{Observational data}, scraped from the web or collected from
sensors, records what happened but not why; the rules that generated
it are unknown, and the same picture or measurement can be consistent
with many underlying mechanisms. \emph{Synthetic data}, produced by a
simulator under a fixed sweep of scene parameters, has a known
generating process but only for the simulator's own template; the
bracket, storm, alloy microstructure, vessel, or patient a downstream
user actually faces was never the scene being swept. The result is a
gap that recurs in every quantitative discipline: a learner that
predicts an outcome cannot say whether the outcome would change if a
physically meaningful parameter were perturbed for \emph{this} case.

This gap is felt acutely in computational science. Learned weather
emulators such as GraphCast~\cite{Lam2023} and Pangu-Weather
\cite{Bi2023} match operational numerical forecasts on in-distribution
skill metrics but inherit the structural assumptions of the reanalysis
they were trained on, and their behaviour under unseen forcings
remains an active concern; large-scale materials discovery from
graph-network predictors trained on density-functional-theory (DFT)
corpora has produced enormous candidate sets but a long tail of false
positives once cross-checked against synthesis or higher-fidelity
simulation~\cite{Merchant2023, Cheetham2024}; structure prediction at
the scale of AlphaFold~\cite{Jumper2021} delivers static structures
with calibrated confidence but does not, on its own, supply the
dynamic, mechanistic, counterfactual data needed for downstream
biophysical reasoning; learned fluid surrogates trained on fixed
Reynolds-number ranges break out of
regime~\cite{PfaffMeshGraphNets2021}; patient-specific cardiac and
cardiovascular models, although mature as forward solvers, remain
bottlenecked by per-case parameter identification and
validation~\cite{Niederer2019}; and medical-image classifiers are
repeatedly shown to exploit scanner artefacts rather than
anatomy~\cite{Geirhos2020}. At the foundation-model layer, the same
pattern recurs: additional web tokens deliver diminishing
returns~\cite{Hoffmann2022, Sambasivan2021, Gadre2023}, and persistent
reasoning failures read as symptoms of data that is correlation-rich
but causation-poor~\cite{Scholkopf2021}. Across these fields the
missing ingredient is the same: data that is anchored on a specific
observation, carries the mechanistic model behind its label, and can
be edited to ask physically meaningful counterfactual questions.

Computational science already maintains the right object for this:
a fully specified forward model with a documented verification-and-validation
record~\cite{Oberkampf2010, ASMEVV10}. Our proposal is to treat that
object not as the \emph{end} of an analysis but as the \emph{data
source} for downstream learning. We call data
produced this way \emph{instrumented data}: every datum is a tuple
that ships with the machinery used to produce it. The world-models
programme~\cite{HaSchmidhuber2018, LeCun2022, Hafner2023} pursues a
related generative stance with learned, implicit, correlation-induced
simulators; instrumented data is the explicit, mechanistic, V\&V-recorded
complement, useful as a substrate for training such models, validating
them, and grounding their counterfactuals.

A recent multi-agent demonstration shows the manufacturing step is
operationally feasible: from a single photograph, agents extract
geometry and material under uncertainty, mesh, solve, verify against
analytical bounds, and produce a code-compliant report autonomously in
minutes~\cite{Wilke2026}. The substrate, not the pipeline, is the
object of this Perspective. The same datum type can be produced by a
cryogenic-electron-microscopy-to-molecular-dynamics (CryoEM$\to$MD)
workflow in structural biology, a
satellite-image-to-regional-climate-model (satellite$\to$RCM) workflow
in geosciences, a microstructure-to-crystal-plasticity-finite-element
(microstructure$\to$CPFE) / DFT workflow in
materials, a particle-image-velocimetry-to-large-eddy-simulation
(PIV$\to$LES) workflow in fluid mechanics, or a
radiology-to-patient-specific finite-element (FE) workflow in medical
imaging.

\textbf{A robustness spectrum, not a binary.} Any instrumentation
pipeline is \emph{more robust} on a problem class when cases sit
inside its validation envelope (interpolative regime), \emph{less
robust} when they sit outside (extrapolative regime). Robustness is a
property of the pipeline-on-class pair, and it conditions every
downstream use (Section~\ref{sec:uses}) and cross-pipeline review
protocol (Section~\ref{sec:produces}).

\textbf{Glossary for the broad reader.} We use a small set of terms
from different communities. \emph{Verification} and \emph{validation}
(V\&V) are the standard checks from computational mechanics that ask,
respectively, whether the equations are solved correctly and whether
they are the right equations for the case~\cite{Oberkampf2010,
ASMEVV10}. The \emph{$\mathrm{do}$-operator} is Pearl's notation for
an intervention, ``what happens if I set parameter $\theta$ to value
$\theta^\star$''~\cite{Pearl2009}; here, doing so means re-running the
solver. \emph{Aleatoric} uncertainty is irreducible (sensor noise,
material variability); \emph{epistemic} uncertainty is reducible by
more information (viewpoint ambiguity, model-form uncertainty).
\emph{Image-to-simulation} (\imgsim{}) refers to a pipeline that
converts a sensor observation into a runnable mechanistic simulation;
it is the running example, not the only realisation. The solver
$\mathcal{S}$ may be a partial-differential-equation (PDE) discretiser
or, for many time-series problems, a far cheaper
ordinary-differential-equation (ODE) integrator or reduced-order
surrogate; the substrate definition does not privilege either. A
\emph{surrogate} is a cheap neural approximation to an expensive
solver; \emph{amortised} surrogate training pays the simulation cost
once so inference is fast. \emph{Interpolative} and \emph{extrapolative}
refer to a case's position relative to a pipeline's validation
envelope, not to numerical interpolation between samples. \emph{Large
language model} (LLM) denotes the transformer-based agent driving
perception, orchestration, and review.

\section{What an instrumented datum is}
\label{sec:produces}

The instrumented datum is defined independently of any specific
pipeline. Let $I$ denote a sensor observation and
$\mathcal{P}: I \mapsto \mathcal{M}$ a V\&V instrumentation pipeline
mapping $I$ to a simulation model
$\mathcal{M} = (\Omega, \sigma, \partial\Omega, u_0, f, \mathcal{S})$:
geometry $\Omega$, governing law $\sigma$, boundary conditions
$\partial\Omega$, initial conditions $u_0$, forcing $f$, and solver
$\mathcal{S}$ (PDE, ODE, multiphysics, or reduced-order). The solver
returns a response $u = \mathcal{S}(\mathcal{M})$. The \imgsim{}
pipeline of~\cite{Wilke2026} is one realisation; CryoEM$\to$MD,
satellite$\to$RCM, microstructure$\to$CPFE, PIV$\to$LES, and
radiology$\to$patient-specific FE workflows are others.

\begin{figure}[!htb]
\centering
\includegraphics[width=0.85\linewidth]{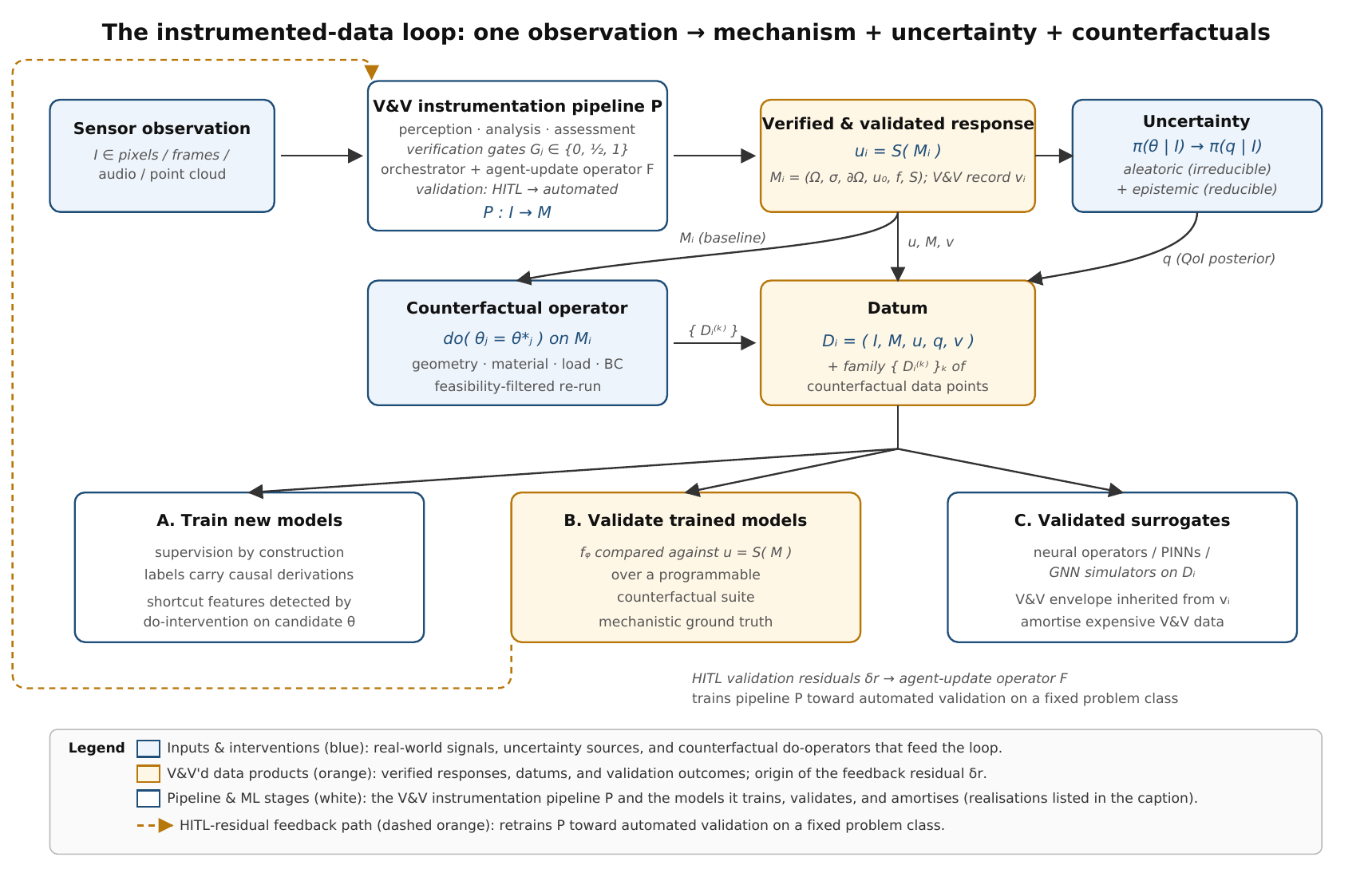}
\caption{The instrumented-data loop. A sensor observation is converted
by a V\&V instrumentation pipeline $\mathcal{P}$ into a verified and
validated response $\mathcal{S}(\mathcal{M}_i)$, with the confounder
bundle $\eta_i$ carried alongside $\mathcal{M}_i$. Realisations of
$\mathcal{P}$ include image-to-simulation pipelines in mechanical
engineering~\cite{Wilke2026}, CryoEM-to-molecular-dynamics workflows
seeded by AlphaFold structures~\cite{Jumper2021}, learned-weather
emulator pipelines on reanalysis-plus-simulation
corpora~\cite{Lam2023, Bi2023},
microstructure-to-crystal-plasticity / DFT
workflows~\cite{Merchant2023}, PIV-to-LES graph-network
surrogates~\cite{PfaffMeshGraphNets2021}, and radiology-to-patient-specific
finite-element workflows~\cite{Niederer2019}. Validation is initially
supplied by a human-in-the-loop (HITL) domain expert and, through the
agent-update operator $F$ trained on HITL residuals, is anticipated
(not yet demonstrated) to migrate toward automated validation on a
fixed problem class. Extraction uncertainty yields a push-forward
$\pi(q\mid I)$ resolved into aleatoric (irreducible) and epistemic
(reducible) components, together with counterfactual interventions
$\mathrm{do}(\theta_j\!=\!\theta_j^\star)$ on either mechanistic or
confounder parameters. The datum $\mathcal{D}_i$ feeds five consumers
(Section~\ref{sec:uses}). Three are shown here as direct downstream
consumers of the loop: training of new models (Use~1), validation of
existing models against mechanistic ground truth (Use~2), and training
of validated surrogate networks that amortise cost (Use~3). Two further
uses sit upstream and beside the loop: fewer-but-richer pretraining
for scientific-reasoning foundation models (Use~4) and on-demand
reasoning tools that an LLM agent invokes at inference time (Use~5).}
\label{fig:loop}
\end{figure}

A datum is the tuple
\begin{equation}
\mathcal{D}_i \;=\; \bigl(\,I_i,\; \mathcal{M}_i,\; \eta_i,\; u_i,\; q_i,\; v_i\,\bigr),
\label{eq:datum}
\end{equation}
with $q_i$ the quantity of interest (stress, drag, temperature, modal
frequency, biomarker concentration, etc.); $v_i$ the combined V\&V
record; and $\eta_i$ the \emph{confounders} carried explicitly outside
$\mathcal{M}_i$: acquisition (viewpoint, illumination, calibration),
environmental (temperature, humidity, drift), and protocol (operator,
equipment) factors that influence $I_i$ but lie outside the mechanistic
model. Naming $\eta_i$ explicitly lets downstream learning condition,
marginalise, or intervene on it rather than absorb it as label noise. $v_i$ carries verification artefacts (mesh convergence, residuals
against analytical bounds, gate outcomes, domain-standard flags) and
validation artefacts (domain-expert sign-off, residuals against
physical measurement, perception-layer calibration history). Unlike a
labelled image, $\mathcal{D}_i$ exposes the causal graph, the
confounders, and the record by which both V\&V questions were
answered. Because $\mathcal{D}_i$ is a typed object carrying its
$v_i$, cross-pipeline review (introduced below) operates on a
structured datum with audit trail, not on free-form agent output.
Figure~\ref{fig:loop} maps the full data-generating loop.

\textbf{Validation: HITL today, automated tomorrow.} Today, validation
is supplied by a human-in-the-loop (HITL): a qualified domain expert
(scientist, clinician, engineer, etc.) signs each report after
inspecting $v_i$~\cite{Wilke2026}. The pipeline of~\cite{Wilke2026}
already specifies an agent-update operator $F$ that consumes expert
residuals $\delta r$ and writes them back as rules into agent memory,
prompts, and gates. Repeated application of $F$ on a fixed problem
class is \emph{anticipated}, though not demonstrated, to converge
toward \emph{automated validation}: the bands
$[\theta_j^{-}, \theta_j^{+}]$ track empirical residuals from physical
measurement, and the marginal supervisory burden per sample shrinks.
Professional sign-off does not disappear; the time to discharge it
does. A pipeline $\mathcal{P}_A$ that has accumulated such a record
on class $A$ can then act as an external reviewer for a sibling
$\mathcal{P}_B$ via a cross-pipeline gate $G^{A \to B}$ on the typed
datum, generalising the single-pipeline gates
of~\cite{Wilke2026}. We treat this
HITL\,$\to$\,automated\,$\to$\,peer trajectory as a working
hypothesis; only the first round of $F$ (pure HITL sign-off, no
automated update yet) is demonstrated in~\cite{Wilke2026}, and the
risks of automating validation are surfaced in
Section~\ref{sec:risks}.

\begin{figure}[!htb]
\centering
\includegraphics[width=0.98\linewidth]{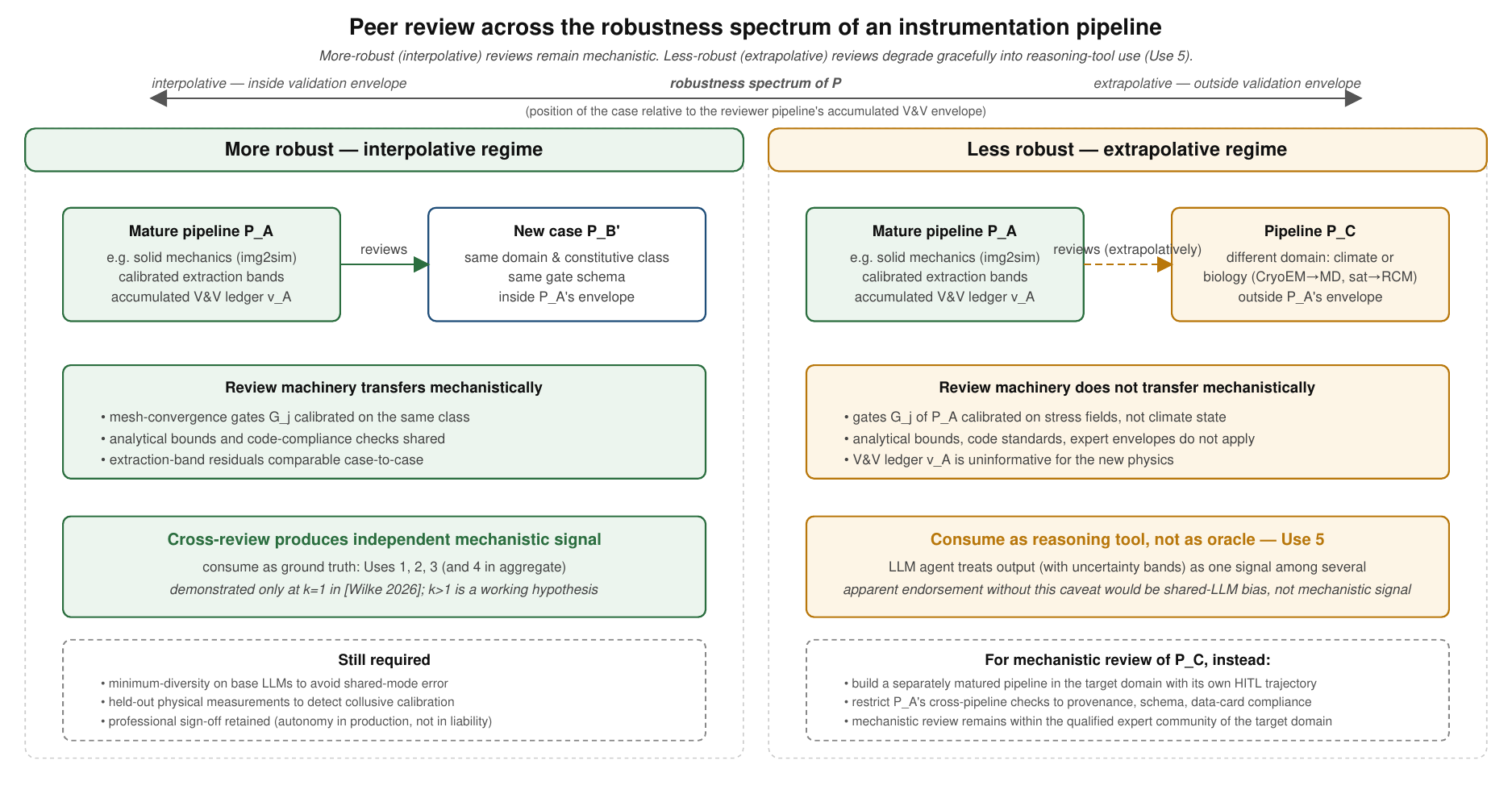}
\caption{When can a mature instrumentation pipeline act as a peer
reviewer? \textbf{Left (interpolative regime, more robust):} a mature
pipeline $\mathcal{P}_A$ (e.g.\ a solid-mechanics \imgsim{} pipeline)
reviewing a new case $\mathcal{P}_B'$ that shares constitutive class,
gates $G_j$, analytical bounds, and code-compliance checks operates
inside $\mathcal{P}_A$'s accumulated validation envelope; the review
produces independent mechanistic signal, subject to a minimum-diversity
requirement on the underlying LLMs, held-out physical measurements, and
retained professional sign-off. \textbf{Right (extrapolative regime,
less robust):} the same $\mathcal{P}_A$ attempting to review a pipeline
$\mathcal{P}_C$ in a different physical domain (e.g.\ climate or
biology) sits outside $\mathcal{P}_A$'s validation envelope; its
gates, bounds, and ledger do not transfer mechanistically, and
apparent endorsement risks being a shared-LLM artefact. Cross-domain
mechanistic review remains the responsibility of a separately matured
pipeline in the target domain and its qualified expert community.
Useful work is still possible in the right-hand regime if the
pipeline output is consumed as a reasoning tool rather than as
ground truth (Section~\ref{sec:uses}, Use~5). Only the interpolative
scenario in the first round of $F$ is demonstrated
in~\cite{Wilke2026}; both regimes beyond the first round are working
hypotheses, with risks itemised in Section~\ref{sec:risks}.}
\label{fig:trajectory}
\end{figure}

The utility of $G^{A \to B}$ is not binary but varies with how far
$\mathcal{P}_B$'s case sits inside $\mathcal{P}_A$'s validation
envelope. Interpolative cross-review (shared constitutive class, gate
schema, review history) is \emph{more robust} and yields independent
mechanistic signal. Extrapolative cross-review (different physics or
expert community) is \emph{less robust}: $\mathcal{P}_A$'s gates no
longer apply mechanistically and apparent endorsement risks being a
shared-LLM artefact. Figure~\ref{fig:trajectory} contrasts the two
ends.

Three properties follow.

\textbf{Causality for the specified case.}
$\mathcal{M}_i \mapsto u_i$ is a known structural causal model in the
sense of Pearl~\cite{Pearl2009}, and $\mathcal{M}_i$ is inferred from
the actual observation $I_i$ rather than chosen from a corpus-time
template. Interventions $\mathrm{do}(\theta_j \!=\! \theta_j^\star)$ on
any parameter $\theta_j \in \mathcal{M}_i \cup \eta_i$ are well defined
and executable: re-run the solver under the new mechanistic or
confounder setting. Distinguishing correlation from causation, the central
difficulty in observational ML, is trivial \emph{within} $\mathcal{M}_i$
because the structural equations are the solver. The case-causality
claim between $I_i$ and $\mathcal{M}_i$ is bounded by the calibration of
the extraction operator $\mathcal{I}$ (Eq.~\ref{eq:extraction}); the
substrate therefore provides \emph{conditional} causality, exact given
$\mathcal{M}_i$ and only as good as extraction otherwise.

\textbf{Counterfactuals.} For each $\mathcal{D}_i$ the pipeline can
emit a family $\{\mathcal{D}_i^{(k)}\}_{k=1}^{K_i}$, where $k$ indexes
counterfactual variants and $K_i$ is the number of admissible
interventions enumerated for case $i$, generated by intervening on
$\mathcal{M}_i$ or on $\eta_i$: \emph{what if this bracket were
stainless?}, \emph{what if the load doubled?}, \emph{what if the
imaging illuminant shifted by 200~K?} Each counterfactual is itself
a verified simulation, not a perturbed pixel.

\textbf{Supervision by construction.} The label $q_i$ is computed from
$\mathcal{M}_i$ and $u_i$ by deterministic post-processing; it needs no
human annotator and inherits the \vv{} record $v_i$. The supervision
signal is auditable.

\section{One image becomes a distribution of simulations}

A single image rarely determines $\mathcal{M}$ uniquely. The
extraction operator returns set-valued, probabilistic, or possibilistic
estimates of each parameter~\cite{Wilke2026}:
\begin{equation}
\mathcal{I}: (I, \text{context}) \longrightarrow
\bigl\{(\hat{\theta}_j, [\theta_j^{-}, \theta_j^{+}], p_j(\theta_j \mid I),
\kappa_j)\bigr\}_{j=1}^{n_\theta},
\label{eq:extraction}
\end{equation}
where $\theta_j$ is the $j$-th parameter of the joint set
$\{\theta_j\}_{j=1}^{n_\theta} = \mathcal{M}_i \cup \eta_i$,
$\hat{\theta}_j$ its point estimate, $[\theta_j^{-}, \theta_j^{+}]$
an interval where no calibration exists, $p_j(\theta_j \mid I)$ a
density where a best estimate does, $\kappa_j$ a categorical
confidence label (e.g.\ a fuzzy class membership) for qualitative
parameters such as material class~\cite{Zadeh1965}, and $n_\theta$
the total number of parameters. Propagating through $\mathcal{S}$
yields a push-forward $\pi(q\mid I)$ over the quantity of interest:
not a confidence bound on a single label but a cloud of admissible,
verified data points. \textbf{Caveat:} in the single-image regime
these bands are agent self-reports, not empirically calibrated
intervals~\cite{Wilke2026}; calibration requires domain-expert ground
truth feeding the update operator (Section~\ref{sec:risks}).
Separating aleatoric from epistemic components of $\pi(q\mid I)$
requires repeated extractions and a calibrated perception layer; this
protocol is itself an open methodological question
(Section~\ref{sec:openq}, item~1).

\section{Five uses, in order of increasing leverage}
\label{sec:uses}

\subsection{Use 1: training data that is causal by construction}

Instrumented data can train downstream models (defect detectors,
materials-property regressors~\cite{Merchant2023}, diagnostic
classifiers, fluid-mechanics drag predictors~\cite{PfaffMeshGraphNets2021},
biomarker estimators built on molecular-dynamics (MD) simulations
seeded by AlphaFold-predicted structures~\cite{Jumper2021})
where every label has a documented mechanistic derivation. This
addresses the most embarrassing failure mode of synthetic data: a
learner exploiting shortcut features absent in
deployment~\cite{Geirhos2020}. When a shortcut is suspected, the
$\mathrm{do}$-operator on the candidate feature settles it.

\subsection{Use 2: automated validation of already-trained models}

A model $f_\phi$ (with $\phi$ its trained parameters, distinct from
the forcing $f$ of $\mathcal{M}$) trained on observational data can
be probed against an instrumented counterfactual suite: intervene on
each physically meaningful parameter, compare $f_\phi$ against
solver-computed truth.
This is \emph{validation} of the learned surrogate against mechanistic
reality, not verification of its numerics, with explicit coverage of
the input manifold. The same audit applies to learned world
models~\cite{HaSchmidhuber2018, Hafner2023}, supplying the calibration
target that the world-models literature currently lacks. The
contribution here is not better models, but credible validation of
existing ones.

\subsection{Use 3: cheap surrogates trained on verified-and-validated data}

Instrumented data \emph{can} be expensive per sample, but cost is
regime-dependent: closed-form benchmarks and low-degree-of-freedom linear problems
are essentially free, while nonlinear contact, large-eddy turbulence,
coupled atmosphere--ocean models, and density-functional materials
simulations dominate the corpus budget. Where the cost is material,
the standard response applies: train \emph{validated} neural
surrogates, whether neural operators~\cite{Li2021, Kovachki2023},
physics-informed networks~\cite{Raissi2019}, or graph network
simulators~\cite{SanchezGonzalez2020, PfaffMeshGraphNets2021}, on the
corpus, and deploy at inference. The same recipe is now standard
practice across fields: learned weather emulators trained on
reanalysis-plus-simulation data~\cite{Lam2023, Bi2023},
materials-property surrogates trained on DFT
corpora~\cite{Merchant2023}, structure-conditioned biomolecular
models building on AlphaFold predictions~\cite{Jumper2021}, and
patient-specific cardiac surrogates trained on finite-element
ensembles~\cite{Niederer2019}. Because each instrumented training
datum carries the \vv{} record $v_i$, the surrogate inherits known
coverage, noise structure, causal interventions, and a validation
envelope; none of these holds for surrogates trained on heterogeneous
observational data. Pay once for V\&V-instrumented data, amortise
across many cheap predictions. In modern terminology, a surrogate
trained this way is a \emph{grounded} world model for its domain: a
learned predictor whose training distribution carries explicit
mechanism, uncertainty, and counterfactual coverage.

\subsection{Use 4 (long-term, speculative, robustness-sensitive): fewer-but-richer pretraining}
\label{sec:use4}

The most leveraged use lies upstream of any one task. Every
$\mathcal{D}_i$ carries five objects rarely present in pretraining
corpora: $\mathcal{M}_i$, $u_i$, $\eta_i$, the aleatoric/epistemic
decomposition of $\pi(q\mid I)$, and the counterfactual family
$\{\mathcal{D}_i^{(k)}\}$. We state the postulate in falsifiable form.
\emph{There exists a task family $\mathcal{T}$ of causal,
counterfactual, and calibration-aware benchmarks (e.g.\ CLadder-style
causal reasoning~\cite{Jin2023CLadder}, counterfactual
visual-question-answering / natural-language-inference
(VQA/NLI)~\cite{Kaushik2020}) and a measurement protocol $\Pi$ at
matched compute under which $N$ instrumented samples match $\rho N$
correlation-only web samples on test loss, with $\rho>1$ growing in
instrumentation depth.} Here $\rho>0$ is the
\emph{informational-density ratio} of instrumented over web samples
on the chosen benchmark. The null $\rho\le 1$ rejects the postulate. Two
mechanisms plausibly drive the gain: \emph{counterfactual contrast}
(minimal contrastive sets isolating one mechanism at a time, known to
improve sample efficiency~\cite{Kaushik2020, Scholkopf2021}) and
\emph{auditable reasoning} (chain-of-thought scored against structural
equations, supplying a process-level reward web text lacks). The
argument transfers cleanly to scientific-reasoning foundation models
(weather, materials, biomolecular, mechanics), where the structural
equations are precisely what $\mathcal{M}_i$ encodes; it does not
transfer to commonsense or long-horizon planning. \textbf{Robustness
condition.} The postulate is most defensible when the corpus is
dominated by more-robust, interpolative samples; as the corpus shifts
toward less-robust, extrapolative samples, ingesting mechanistic
``ground truth'' as pretraining supervision risks baking extrapolation
errors in at scale. Use~5 is the safer consumption mode there.
Measuring $\rho$ on $\mathcal{T}$ is the open question of
Section~\ref{sec:openq}.

\subsection{Use 5 (near-term, robustness-tolerant): on-demand reasoning tools for LLM agents}
\label{sec:use5}

Uses 1--3 consume instrumented data as ground truth; Use 4 as
pretraining supervision; Use 5 as a callable \emph{reasoning tool} an
LLM agent invokes at inference time. The pipeline $\mathcal{P}$ with
its push-forward $\pi(q\mid I)$ and operator
$\mathrm{do}(\theta_j\!=\!\theta_j^\star)$ becomes a tool the agent
queries to test hypotheses, run order-of-magnitude ``what if''
questions, or check chain-of-thought against the physics. The agent
treats the response as one signal among several, with explicit
uncertainty bands; residual extraction, solver, and extrapolation
errors are absorbed by the agent's downstream uncertainty handling
rather than propagated as deployment failures.

This is the use case in which the robustness spectrum matters
\emph{least} in the following precise sense. Uses~1--4 demand
\emph{quantitative accuracy}: the pipeline must return numbers the
downstream consumer can trust as ground truth (labels, validation
oracle, surrogate target, pretraining supervision). Use~5 demands
only \emph{qualitative accuracy}: the pipeline must return the
correct direction of an effect (does $q$ increase or decrease when
$\theta_j$ is perturbed?), the right order of magnitude, and a
sensible monotonicity or scaling, rather than a tightly calibrated
absolute value. In the more-robust regime, the tool delivers both;
in the less-robust regime, it can still deliver the qualitative
signal, and that is enough for an LLM agent doing comparative
reasoning. The pipeline only has to be accurate enough to be
\emph{sensible about trends}; absolute precision becomes optional.
This is why Use~5 remains operationally useful in the extrapolative
regime where Uses~2 and~4 are not: trend-accurate but
magnitude-uncertain output is a useful tool signal, whereas the same
output consumed as ground truth or pretraining oracle would be a
source of confident but wrong numbers.

\section{A third data substrate}

Table~\ref{tab:substrates} states the trichotomy. Each column is the
answer to one question a downstream learner cares about: does each
datum carry its mechanism; is the mechanism specific to the case the
user faces; can a physically meaningful intervention be executed; is
there an auditable V\&V record?

\begin{table}[!htb]
\centering
\footnotesize
\caption{Three data substrates for scientific machine learning,
distinguished by case-specific mechanism, executable counterfactual,
and V\&V record per datum.}
\label{tab:substrates}
\setlength{\tabcolsep}{4pt}
\begin{tabular*}{\textwidth}{@{\extracolsep{\fill}}lcccc@{}}
\toprule
\textbf{Substrate} & \textbf{Mechanism?} & \textbf{Case-specific?} & \textbf{Counterfactual?} & \textbf{V\&V record?}\\
\midrule
Observational           & No  & Yes & No            & No\\
Template synthetic      & Yes & No  & Template only & Sometimes\\
\textbf{Instrumented}   & \textbf{Yes} & \textbf{Yes} & \textbf{Yes} & \textbf{Yes}\\
\bottomrule
\end{tabular*}
\end{table}

The instrumented row collects properties the other two never
simultaneously satisfy. Standard objections to synthetic data are
addressed item by item. \emph{Domain gap:} the simulation is
conditioned on a real observation, not a procedural scene; confounders
are named in $\eta_i$ rather than absorbed as noise. \emph{Coverage:}
the push-forward $\pi(q\!\mid\!I)$ and $\mathrm{do}(\cdot)$ make
coverage a property of the intervention policy, not of the dataset.
\emph{Circularity:} the V\&V record bounds solver bias per sample, so
circularity becomes auditable rather than aggregate~\cite{Wilke2026,
Oberkampf2010}.

Instrumented data is not a replacement for observational data but a
third category, shipped with the machinery used to produce it. The
same trichotomy applies whether the datum was produced by an
\imgsim{} pipeline~\cite{Wilke2026}, a CryoEM$\to$MD workflow seeded
by AlphaFold structures~\cite{Jumper2021}, a learned-weather emulator
on reanalysis-plus-simulation~\cite{Lam2023}, a
microstructure$\to$CPFE workflow on DFT constitutive
laws~\cite{Merchant2023}, a PIV$\to$LES graph-network
surrogate~\cite{PfaffMeshGraphNets2021}, or a
radiology$\to$patient-specific FE workflow~\cite{Niederer2019}.

\section{Risks, limits, and what must still be earned}
\label{sec:risks}

Instrumented data is not automatically trustworthy. Six risks must be
confronted before scale. \textbf{(i) Perception calibration:}
extraction bands are agent self-reports until empirical coverage is
verified against physical measurement. \textbf{(ii) Solver fidelity
bounds the substrate:} plasticity, fracture, turbulence, and
multi-physics coupling have regimes where solver error dominates
extraction error, and $v_i$ must surface this.
\textbf{(iii) Professional oversight is not optional:} a qualified
domain expert signs off, as in conventional V\&V.
\textbf{(iv) Counterfactual realism:} not every
$\mathrm{do}(\theta_j)$ on $\mathcal{M}_i$ or $\eta_i$ yields a
physically realisable configuration; a feasibility filter is
mandatory. \textbf{(v) Distribution shift:} instrumented data narrows
the deployment gap by anchoring on real observations; it does not
eliminate it. \textbf{(vi) Robustness mismatch:} consuming an
extrapolative pipeline as if it were interpolative (using it as a
Use~2 or Use~4 oracle rather than a Use~5 reasoning tool) is the
most likely way to convert honest mechanistic uncertainty into a
confident deployment error.

\section{Methodological questions for the community}
\label{sec:openq}

Nine open questions will determine whether instrumented data matures
into a recognised substrate for scientific machine learning.

\begin{enumerate}[leftmargin=1.4em,itemsep=0.2em]
\item \textbf{Calibration protocols.} How much physical ground truth
is needed before agent-reported bands can be promoted to calibrated
intervals on a class? Conformal prediction~\cite{Vovk2005,
AngelopoulosBates2023} over extraction--measurement residuals is a
natural candidate.
\item \textbf{Counterfactual coverage metrics.} What is the analogue
of dataset coverage in $\theta$-space, weighted by causal relevance
to the downstream quantity of interest?
\item \textbf{Verification of the verifier.} Quality gates are
LLM-driven; auditing their calibration requires adversarial probing,
inter-agent disagreement, and held-out physical measurements.
\item \textbf{Provenance and licensing.} A standardised data card
extending Datasheets for Datasets~\cite{Gebru2021} is needed to carry
the inheritance from image, solver, gates, and reviewer.
\item \textbf{Cost--accuracy frontiers.} Open benchmarks pairing
instrumented corpora with surrogates are needed to quantify when
surrogate training amortises the substrate's cost.
\item \textbf{Robustness-conditional peer-review protocols.}
The community needs an empirical map of how cross-pipeline review
degrades from interpolative to extrapolative regimes, a
minimum-diversity requirement on participating LLMs, and a
demarcation rule restricting extrapolative checks to provenance,
schema, and data-card compliance.
\item \textbf{Quantifying the fewer-but-richer postulate.}
Paired benchmarks at equal compute that pin down the
informational-density ratio $\rho$ in Section~\ref{sec:use4} as a
function of task type and instrumentation depth. Without this
measurement, Use~4 remains plausible but unfalsified.
\item \textbf{HITL\,$\to$\,automated validation as a risk surface.}
Three failure modes need protocols: \emph{shared-mode error}
(diversity across base LLMs), \emph{collusive calibration}
(held-out physical measurements), and \emph{autonomy in production is
not autonomy in liability} (sign-off retained at federation level).
\item \textbf{Reasoning-tool evaluation for less-robust regimes.}
Tool-use benchmarks paired with V\&V-instrumented tools, with
robustness labels per call, are needed to score Use~5 agents that
weight uncertainty bands and downweight extrapolative calls.
\end{enumerate}

\section{Conclusion}

The hard problem in scientific machine learning is no longer fitting
parameters but producing data whose causal structure is known.
V\&V-instrumented pipelines, of which image-to-simulation is one
realisation, deliver exactly that: each datum is an observation, a
mechanistic model, a confounder bundle, a computed response, an
aleatoric/epistemic uncertainty, and an executable counterfactual
family. More expensive per sample than scraped data; more causal,
verifiable, and validatable than the alternatives. Its value is
conditional on where the pipeline sits on the robustness spectrum:
the more interpolative, the stronger as quantitative ground truth
(Uses~1--3) and pretraining oracle (Use~4); the more extrapolative,
the more its consumer must be an LLM agent treating it as a
trend-level reasoning tool (Use~5), where qualitative accuracy on
direction, sign, and order of magnitude is sufficient and absolute
precision is not required. The work to be done is calibration, coverage, governance, a
clean measurement of $\rho$, and tool-use benchmarks that respect
robustness.

\section*{Code and data availability}
The base multi-agent \imgsim{} system is described in~\cite{Wilke2026};
its orchestrator prompt and demonstration assets are released with that
work's supplementary material. No new datasets or models are introduced in
this position paper.

\section*{Competing interests}
The author declares no competing interests.


\end{document}